\documentclass{article}

\usepackage{arxiv}

\usepackage[utf8]{inputenc} % allow utf-8 input
\usepackage[T1]{fontenc}    % use 8-bit T1 fonts
\usepackage{hyperref}       % hyperlinks
\usepackage{url}            % simple URL typesetting
\usepackage{booktabs}       % professional-quality tables
\usepackage{amsfonts}       % blackboard math symbols
\usepackage{nicefrac}       % compact symbols for 1/2, etc.
\usepackage{microtype}      % microtypography
\usepackage{lipsum}
\usepackage{graphicx}
\usepackage{cite}
\usepackage{multicol}

\title{LiDAR Based Detection and Classification of Pedestrians and Vehicles Using Machine Learning Methods}

\author{
  Farzad Shafiei~Dizaji\\
  Engineering Systems and Environment\\
  University of Virginia\\
  Charlottesville, VA 22904 \\
  \texttt{ffs5da@virginia.edu} \\
}

\begin{document}
\maketitle

\begin{abstract}
The goal of this paper is to classify objects mapped by LiDAR sensor into different classes such as vehicles, pedestrians and bikers. Utilizing a LiDAR-based object detector and Neural Networks-based classifier, a novel real-time object detection is presented essentially with respect to aid self-driving vehicles in recognizing and classifying other objects encountered in the course of driving and proceed accordingly. We discuss our work using machine learning methods to tackle a common high-level problem found in machine learning applications for self-driving cars: the classification of pointcloud data obtained from a 3D LiDAR sensor. 
\end{abstract}

% keywords can be removed

\begin{multicols}{2}
\section{Introduction}
LiDAR is a surveying approach that measures distance to a target by illuminating that target with a pulsed laser light and measuring the reflected pulses with a sensor. Differences in laser return times and wavelength can then be used to make digital 3D representations of the target \cite{premebida2007lidar}.

There has been considerable work on object classification for autonomous vehicle navigation. Most of the works use camera vision to capture 2D images of vehicles and other vulnerable road users such as pedestrians and bicyclists. However, camera vision alone may not be able to provide important depth information to detect and track objects with the level of reliability needed for safe driving. LiDAR enables generation of 3D images from a single sweep. Multiple sweeps can be used to generate information about velocities and distances. For greater accuracy, we may use multi-sensor fused data, which is already used for traffic light detection \cite{fairfield2011traffic}. In our project, we intend to train the data using Neural Networks after any required reprocessing or feature extraction. Neural Networks use the processing of the brain as a basis to develop algorithms that can be used to model complex patterns or prediction problems \cite{lawrence1997face}. The basic network architecture of neural networks has an input layer, hidden layer(s) and an output layer. The hidden layers are used as ‘filters’ that make the network faster and efficient by identifying the important patterns of input that must be passed on to the next layers. Neural networks have the ability to learn from initial inputs and their relationships, generalize the model and predict on unseen data. Our goal is to create a system that can efficiently and accurately predict and classify ahead of time the objects in its surroundings, so as to aid the self-driving car in maneuvering through safely.

The LiDAR sensor data is present as a dense three-dimensional point cloud. Before it can be used for classification, a lot of pre-processing needs to be done to refine the data, cluster the point cloud and reduce the number of clusters. To begin with, the point cloud frame data is extracted, following which ground filtering is carried out in an effort to facilitate the clustering process. The Random Sample Consensus (RANSAC) algorithm is used for the ground filtering process. RANSAC operates by producing best fit linear or quadratic planes by observing objects in the environments as outliers. From there, we can isolate grounded points by thresholding points in the z-axis. Next, “Density Based Spatial Clustering of Application with Noise” (DBSCAN) algorithm is used for clustering. It is a density-based algorithm which is simple and fast and does not require any initialization and allows for an unlimited number of clusters. The approach produces clusters based on euclidean distance criteria and a minimum number of samples. Following clustering, valid and usable features are extracted from the clusters. This generates useful candidates that can be fed as input data to machine learning algorithms for classification via Support Vector Machines and Artificial Neural Networks.

As autonomous vehicles advance towards handling realistic road traffic, they face street scenarios where the dynamics of other traffic participants must be considered explicitly \cite{levinson2011towards}. LiDAR enables extensive 3D mapping of its environment, which can be used to navigate a self-driving car or robot predictably and safely. By using LiDAR sensor data, it is possible to generate position data for objects surrounding the car ahead of time \cite{premebida2007lidar, kidono2011pedestrian}. This becomes essential in achieving full autonomy in self driving cars, especially in areas with unpredictable traffic situations. In this project, we seek to enable quick classification of LiDAR sensor data into vehicles, pedestrians, bikers, etc. to enable a self-driving car to plan its trajectory accordingly.

The paper is presented as follows: we formally introduce our problem in section 2 and provide some background in section 3. We then discuss the KITTI dataset and the preprocessing needed to perform feature extraction and classification operations in sections 4, 5. We then describe our classification architecture and our experimental results in 6. Finally, we conclude with information on our software and hardware and some final remarks in section 7 and section 8.

\section{Problem Definition}
Consider an autonomous agent in an unknown environment with no extra information about the potential
shape, dynamics, or behaviors of the environment. Our agent seeks to identify objects in this environment
given by structural data captured in the form of pointclouds which are defined as tuples containing an x,
y, z coordinate and an intensity value. Given a pointcloud O, there exists some subset of O that includes
points belonging to a specified class. In our work, we consider three classes: pedestrians, cars, and cyclists.
We desire then to correctly label each point so that it belongs to a subset Oped, Ocar, Ocyclist, and Oignored
(to indicate points which fit no class but can be safely ignored such as walls or trees).

\section{Background}
In existing literature, we can identify a plethora of work with a large surge since the 2007 DARPA Urban
challenge. Dr. Sebastian Thrun at Stanford University has been a pivotal investigator of this work, producing
a great wealth of research on object tracking using shape, color, and motion \cite{held2014combining}, object tracking with semisupervised learning \cite{teichman2012tracking}, and segmentation of RGBD data \cite{teichman2013learning}. From these contributions, a standard method
for handling 3D pointcloud data has emerged. First, if there are multiple sources of LiDAR data, the
pointclouds are fused according to some registration technique. In many cases, this is conducted via Iterative
Closest Point (ICP), an optimization algorithm which seeks to minimize the error between corresponding
points in two pointclouds. From there, the data is segmented into objects of interest. Segmentation draws a
parallel in machine learning terminology in clustering as its major goal is to produce groupings of interesting
data points which may be used for tracking and classification. Segmentation is an active research topic but is
commonly performed using one of a variety of techniques including DBSCAN, neural network classification,
and geometric fitting. Being real-time is one of the most important characteristics of an algorithm in this field. But there are numerous factors that can impact this capability such as path planning. A rapid path planning algorithm is needed that can generate the shortest path from an initial point to a goal point to gather the essential information from the surrounding environment. Recently an exact geometry based path planning algorithm has been presented in \cite{jafarzadeh2014new, jafarzadeh2018exact
} which has $O(nn'^2)$ running time where $n$ shows the number of the vertices and $n'$ is the number of effective polygons which can be convex or non-convex. This algorithm has shown its capabilities in finding a collision-free and shortest path among a group of polygons in a very short time. Also, in order to produce real-time speeds, the technical space becomes more
limited and algorithms like DBSCAN are used less due to their long processing times.
Finally, after the objects have been segmented, the data can be evaluated in parallel for tracking and
classification. Tracking extracts relevant state information on the cluster subject such as velocity, volume,
position, reachable space, etc. This is commonly performed using an Extended Kalman Filter that attempts
to figure out the corresponding A, B, C, and D matrices in state-space convention that dictate the dynamics
of the object. Classification attempts to take relevant information about the object and identify what
type of object it is, such as a cyclist, pedestrian or a car. Generally, these do not need the information
extracted from tracking but new research is utilizing some of the motion characteristics of an object to help
classification. This is commonly performed with some feature extraction and then either artificial neural
networks or support vector machines. In this work, we choose to show our results in testing a SVM, ANN,
and a decision tree.
\end{multicols}

\begin{figure}[h]
  \centering
  \includegraphics[width=0.5\textheight]{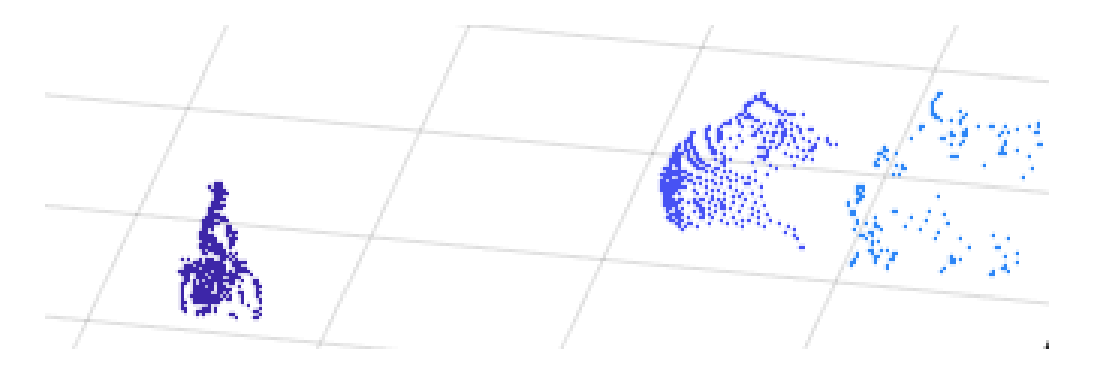}\\
  \caption{Sample label in KITTI dataset with two cars and a cyclist}
  \label{sampleSet}
\end{figure}

\begin{multicols}{2}
\section{Dataset}
We use the KITTI dataset \cite{geiger2012we}, to train our algorithms. The KITTI dataset, released in 2012 contains several
hours of LiDAR footage with object labels for various classes including our desired pedestrian, cyclist, and
car classes. The dataset is also widely recognized as an excellent method for standardized testing as the
dataset appears in many works of literature.  In Figure ~\ref{sampleSet}, a sample set of labeled points is shown clustered by color. In the image, a cyclist and two vehicles can be somewhat clearly identified.

\subsection{Data and Preprocessing}

We currently plan to use the KITTI Vision dataset, which is publicly available and contains six hours worth of frame-by-frame traffic footage. The data contains both camera images and LiDAR point maps, as well as tags for which images contain which objects \cite{geiger2013vision}. We hope to use these features and compatibility with existing software to which we have access to make it simpler to classify the data.

Depending on the format and labeling of the data, we may need to perform our own registration and segmentation on the data. If given multiple streams of LiDAR data, we may need to fuse these different sources from their different perspectives to achieve better results. This can be performed easily using the open source PCL library and its variety of registration algorithms such as iterative closest point or feature matching through the use of SIFT features \cite{rusu2011point}. Segmentation takes the raw data and finds continuities, grouping densities of data into classifiable objects. The segmented data can then be used for machine learning applications \cite{grilli2017review}.

\subsection{Algorithms}

The algorithms that we intend to use for classification in our project are Neural Networks and Support Vector Machines. Neural Networks are greatly used in pattern recognition and classification problems, owing to their ability to take in a lot of inputs, and process them to infer complex, non-linear relationships \cite{lawrence1997face}. Support Vector Machines are classifiers that use labeled training data to generate an optimal hyper-plane that can be used to classify unseen data \cite{sang2008least,cristianini2000introduction}.

\section{Pre-processing}

Since the KITTI datset presents completely raw data, we must utilize a way to filter out ground data and to
cluster the objects appropriately. These are both sources of error in our work but also exist as new potential
avenues of future research and have appropriately large amounts of literature. In our case, we chose to use a
method of ground filtering produced by an author at Beijing Normal University known as Cloth Simulation
Filtering (CSF) \cite{zhang2016easy} and the Mean Shift algorithm for clustering.

\subsection{CSF Ground Filtering}

CSF ground filtration relies on a simulated model of a cloth commonly utilized by computer graphics. The
cloth is assigned a certain rigidity and is laid on an inverted version of the point-cloud. Due to the natural
behaviors of the cloth, rolling hills in the cloth are captured while sharp changes indicating objects are
ignored. The method requires some tuning but yielded excellent results. As a side note, assigning extreme
rigidity to the cloth essentially forces this algorithm to become a plane fitting algorithm.

\subsection{Mean-Shift Clustering}

\begin{figure*}[t]
 \centering
  \includegraphics[width=0.36\textwidth, height=0.246\textheight]{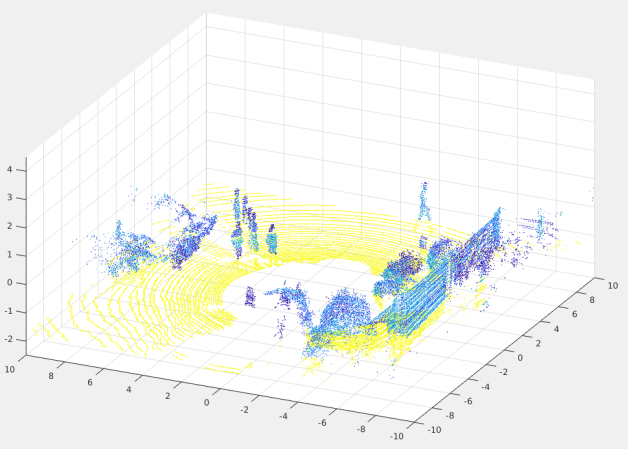}
  \includegraphics[width=0.36\textwidth, height=0.246\textheight]{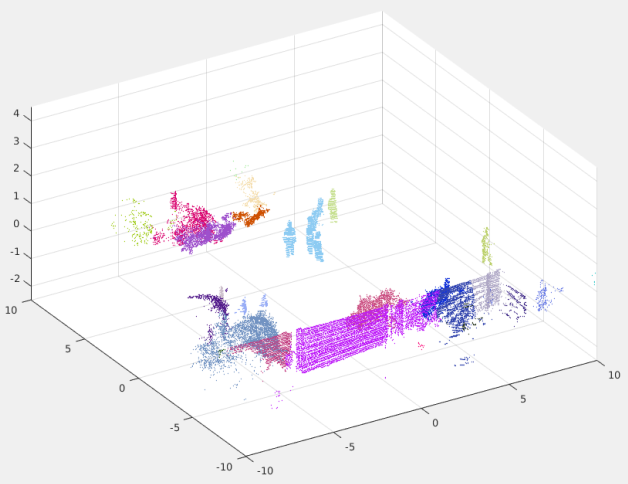}\\
  \caption{Ground Filtering via CSF - Ground Points are marked in Yellow \& Clustering via Mean-Shift - Clusters are marked by different colors}
  \label{GroundFilter}
\end{figure*}

Mean-shift clustering is an extension of Kernel Density Estimation (KDE), a technique used to estimate an underlying probability density function of a specific set of data points. Mean-shift clustering uses a Gaussian kernel in KDE to evaluate a data-set and observe strong groupings in the data. Each point then is iteratively moved to the closest grouping of data, creating natural clusters based on densities of points. A nice advantage of this approach is that the approach does not require knowledge of the number of clusters to identify which is an important for our work. The approach however, also suffers from speed in larger frames of point-cloud data although for our approach the speed was sufficient (Figure ~\ref{GroundFilter}).
\end{multicols}

\begin{multicols}{2}

\section{Pipeline} 

Features were selected based on features found in existing literature. In many works we found common reference to the use of eigenvalues or eigenvectors to indicate the directionality of a pointcloud. For example, a pedestrian would have a large eigenvalue in the upward direction since the characteristic shape of a pedestrian is thinner than a cyclist or a vehicle. In the end, we selected five features found in literature which were the following. The eigenvalues of the covariance matrix for each cluster in x, y, and z directions, the total volume of the object, and the variance in intensity of the cluster. These features were extracted from 23 of the testing data clusters identified by the object labels for training while the remaining 13 were used for validation.

Because the KITTI dataset was very large, we found that it was extremely difficult to utilize k-fold cross validation for our approach due to the limited amount of data we had available and the limited amount of computational time we had to conduct an adequately size k-fold cross validation. Since a single extraction and training sequence took several hours due to the ground filtering and clustering pre-processing, we decided that breaking the dataset into folds was possible but entirely too time consuming given our current hardware. However, we discuss here that model selection approach is easily doable with enough time and strong enough computing power. In total, we were able to extract features from 16,314 car samples, 1,016 cyclist samples, and 2,918 pedestrian samples.

\begin{figure}[h]
  \centering
  \includegraphics[width=0.5\textheight]{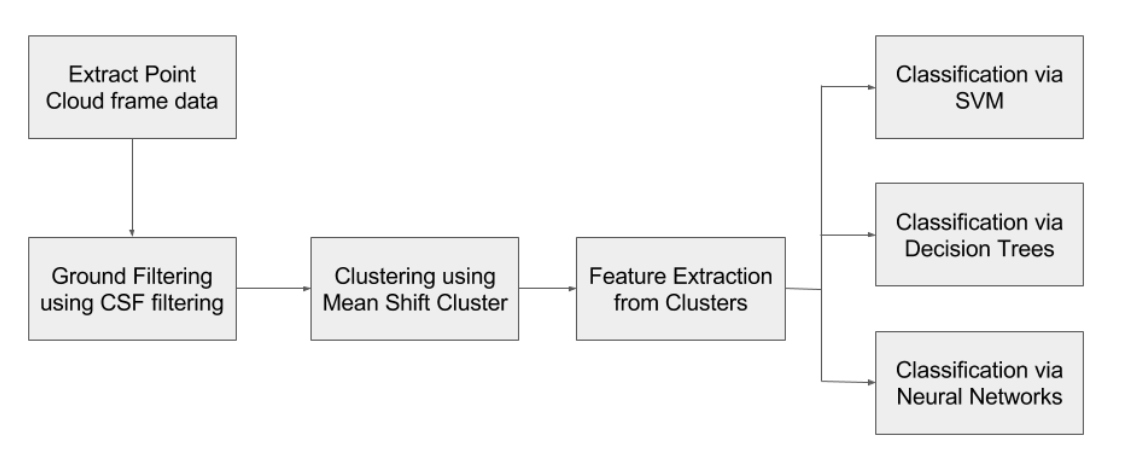}\\
  \caption{Experimental Architecture}
  \label{flowchart}
\end{figure}

Our experimental design utilized the previous section’s work to train and validate a support vector machine and a decision tree. The approach is graphically shown in Figure ~\ref{flowchart}. After conducting the pre-processing stage of the pointcloud data, we fit/train our algorithms using the features extracted from the clusters of each frame of the training set. Then, the data is used to classify the validation data for testing purposes to identify quality.
\end{multicols}

\begin{figure}[t]
  \centering
  \includegraphics[width=0.75\textheight]{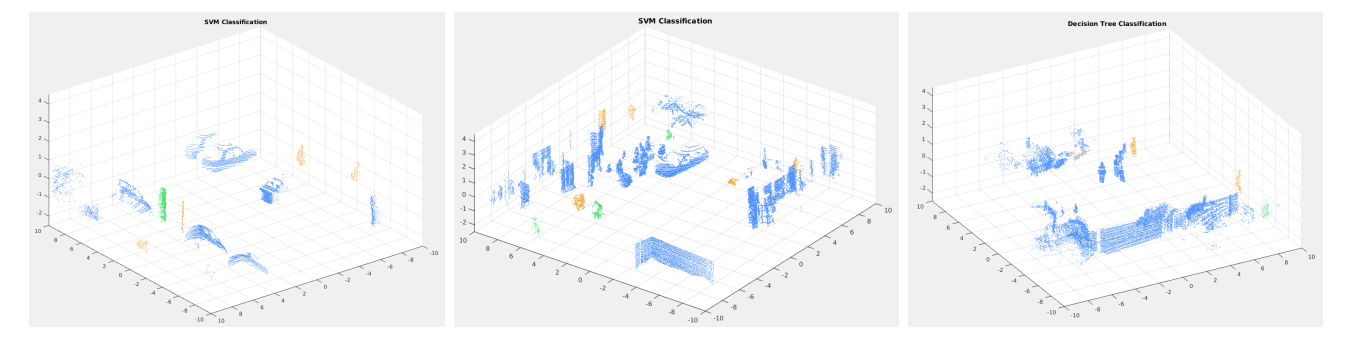}\\
  \caption{Experimental Architecture}
  \label{triple}
\end{figure}

\begin{multicols}{2}

\section{Results}

We present the results of our work with the following metrics. Due to the unique nature of our dataset, we
present the data based on the percentages of each frame that was correctly annotated. In general, we found that our approach yielded a highly noisy but accurate classification methodology. The metrics are presented as the total percentage of data points correctly classified and the total percentage of the labeled data points
correctly classified. The reasoning behind these differing metrics is due to the extreme density of points per frame. Since a significant portion of points belong to no such class, such as walls or trees, the first metric shows how the algorithm should theoretically only classify points in the frame that are of interest. The second metric shows that of the points that are actually of interest, how well does the algorithm perform. For the decision tree and support vector machines, we yielded an overwhelming number of false positives but excellent accuracy. In other words, we found that both approaches yielded high recall but low precision. Our results are presented in the following (Figure~\ref{triple}).

\begin{figure}[t]
  \centering
  \includegraphics[width=0.3\textheight]{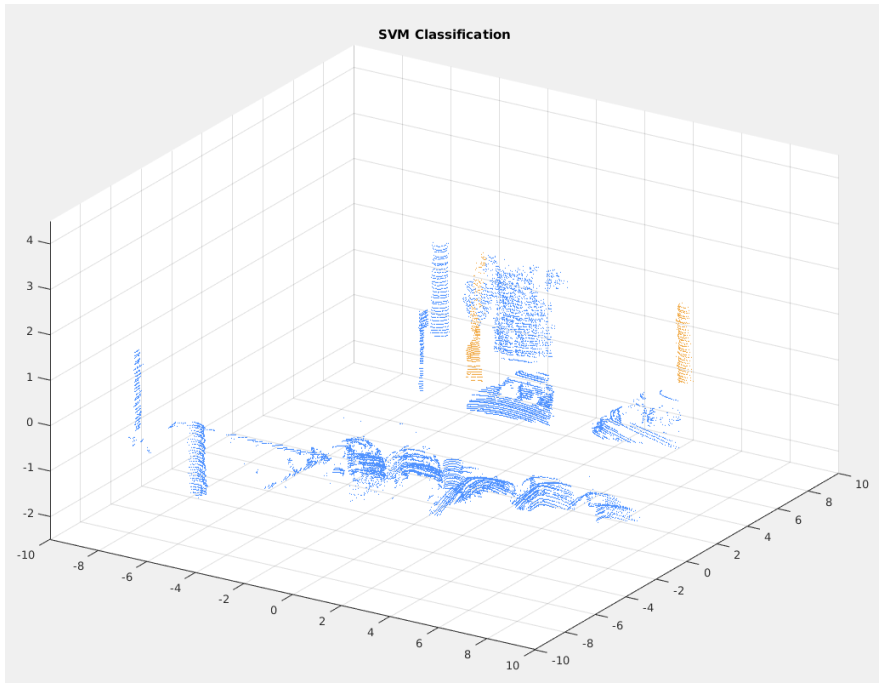}\\
  \caption{Misclassification Examples}
  \label{Misclassification}
\end{figure}

There were a few large sources of error that were easy to understand after going through testing. The first was that we identified objects as ignorable when our classifier had a low prediction score, for our case we used below 90 percent. However, without specifying a class for do not cares, we found that most time our algorithm would just attempt to classify everything with a high margin. This resulted in extremely noisy data and produced incredibly difficult results to understand. Another source of error were for pedestrians and pole shaped objects. Tree trunks, signs, and telephone poles were often misclassified as pedestrians due to their similar eigenvalues and volumes. In general, it would be safe to say that even the intensity variances were comparable. Finally, a final source of error was that our clustering algorithm struggled with objects that were beyond a certain size (Figure ~\ref{Misclassification}). When tuning the clustering parameters, it became difficult to select a bandwidth that was thin enough that close objects were correctly segmented while wide enough that larger objects were not segmented into several components. In future attempts, we could introduce a new layer of pre-processing that may attempt to fuse multiple nearby clusters if there were enough continuous points (Table \ref{accuracy}).

Multi-layer neural networks which are sometimes referred to as perceptrons, are simple models of several connected neurons similar to what is seen in the natural neural networks of animals. The main objective of building these artificial models of brain has been to design systems which can show some learning capabilities like the natural brain \cite{joghataie2009nonlinear,joghataie2012neural}. We also attempted to classify LiDAR data using neural networks. We used MLFFNNs to build our model and will use cross-validation to check the fit of the model (Table ~\ref{table:1}). After studying the dataset, it is clear that the data is skewed and will require proper resampling to train the model effectively. Our plan is to resample the data such that we have a balanced dataset, train our model and check the fit of our model for solving the problem and to choose the best parameters to use in our model. Using the 5 previously mentioned features, we chose only to classify pedestrians and vehicles with deep perceptrons.

\end{multicols}

\begin{table}[h]
 \caption{Number of states based on team parameters}
  \centering
  \begin{tabular}{lll}
    \toprule
    \multicolumn{3}{c}{Accuracy}                   \\
    \cmidrule(r){2-3}
    Classifier     & Total Frame      & Total Cluster \\
    \midrule
    SVM  & 0.04  & 0.638     \\
    DT   & 0.04  & 0.676     \\
    \bottomrule
  \end{tabular}
  \label{accuracy}
\end{table}

\begin{multicols}{2}

According to the data, Class 1 (pedestrian) is under-represented and accounts for only 3 \% of the whole dataset. If we train our model using this dataset, the model will be inefficient and it will be trained to predict only Class 0 because it will not have sufficient training data. One of the other problems of using this skewed dataset to train our model is that since Class 1 is under-represented, the model will assume that it is a rare case and will try to predict positive due to the lack of training data. We may get a high accuracy when we test our model but we should not be confused by this because our dataset does not have a balanced test data. Hence, we have to rely on the recall which relies on TP and FP. In cases where we have skewed data, adding additional data of the under-represented feature (over-sampling) is an option. Since we don’t have that option we have to resort to under-sampling. Under-sampling of the dataset involves keeping all our under-represented data (Class 1) while adding the same number of features of Class 0 (cars) to create a new dataset comprising of an equal representation from both classes. So, after applying Under-sampling method, we have to train/test our model using 2032 rows of data. It is not much but it will have to suffice. Our approach was conducted via MATLAB, sci-kit Learn, and Keras using a standard i7 Intel CPU. Future implementations and approaches could be heavily improved through the use of a GPU and parallel code as the Mean-Shift algorithm is extremely easy to run in parallel.
 
\section{Conclusion}

great deal of pre-processing to filter ground data and extract important features for our classifiers. These elements greatly impacted our approach speed, presenting nowhere near real-time speeds. This is reflected in existing literature as new literature is beginning to move away from SVM and feature-extraction methods and move towards comprehensive neural network based approaches which simultaneously perform feature extraction and classification extremely quickly. These approaches are also naturally useful since they can be leveraged for all aspects of the processing pipeline from classification to segmentation for clustering and for ground filtering. However, as neural networks become more prevalent in self-driving literature an important question arises in how these methodologies work. Many researchers criticize ANN’s for their black-box nature, being able to produce amazing results with no knowledge of their internal workings. This has an extreme relevance in our future culture as future issues that arise with neural networks may be far more difficult to resolve and may lead to catastrophic failures in autonomous systems that rely on their capabilities.

\end{multicols}

\begin{table}[h]
\caption{Table to test captions and labels}
\centering
\begin{tabular}{||c c c c||} 
 \hline
 Layer(type) & Activation Function & Output Shape & Parameters \\ [0.5ex] 
 \hline\hline
 Dense1 & Dense3   & (None, 200) & 1200 \\ 
 Dropout1 & n/a   & (None, 200) & 0 \\
 Dense2 & relu   & (None, 200) & 40200 \\
 Dropout2 & n/a   & (None, 200) & 0 \\
 Dense3 & Sigmoid    &  (None, 1) & 201 \\ [1ex] 
 \hline
\end{tabular}
\label{table:1}
\end{table}

\begin{table}[h]
 \caption{Loss}
  \centering
  \begin{tabular}{lllll}
    \toprule
    Architectures & Training Acc & Training Loss & Validation Acc & Validation Loss \\
    \midrule
    Dense Layer 2   & 0.93  & 0.16 & 0.92 & 0.17  \\
    Dense Layer 3    & 0.95  & 0.16 & 0.95 & 0.17     \\
    Dense Layer 4  & 0.95 & 0.13 & 0.95 & 0.14\\
    \bottomrule
  \end{tabular}
  \label{statesNum}
\end{table}

\begin{multicols}{2}

\bibliographystyle{IEEEtran}
\bibliography{references}

\end{multicols}

\end{document}